\documentclass{article}

\usepackage{arxiv}

\usepackage[utf8]{inputenc} % allow utf-8 input
\usepackage[T1]{fontenc}    % use 8-bit T1 fonts
\usepackage{hyperref}       % hyperlinks
\usepackage{url}            % simple URL typesetting
\usepackage{booktabs}       % professional-quality tables
\usepackage{amsfonts}       % blackboard math symbols
\usepackage{amsmath}
\DeclareMathOperator{\Var}{Var}  % variance
\DeclareMathOperator{\EX}{\mathbb{E}}  % expected value
\usepackage{bm}
\usepackage{nicefrac}       % compact symbols for 1/2, etc.
\usepackage{microtype}      % microtypography
\usepackage{cleveref}       % smart cross-referencing
\usepackage{lipsum}         % Can be removed after putting your text content
\usepackage{graphicx}
\graphicspath{{./}}
\usepackage{natbib}
\usepackage{doi}
\usepackage{subcaption}
\usepackage{siunitx}
\usepackage{arydshln}		% Checkmark
\usepackage{makecell}		% newline in table header

\title{Active Learning to Guide Labeling Efforts for Question Difficulty Estimation}

\date{}

\author{ \href{https://orcid.org/0000-0001-9107-5646}{\includegraphics[scale=0.06]{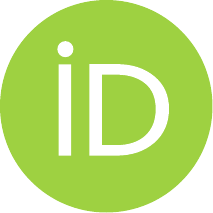}\hspace{1mm}Arthur Thuy}\thanks{Corresponding author} \\
	Ghent University\\
	CVAMO Core Lab, Flanders Make\\
	\texttt{arthur.thuy@ugent.be} \\
	\And
	\href{https://orcid.org/0000-0002-0952-9213}{\includegraphics[scale=0.06]{orcid.pdf}\hspace{1mm}Ekaterina Loginova} \\
	Dedalus Healthcare\\
	\texttt{ekaterina.d.loginova@gmail.com} \\
	\And
	\href{https://orcid.org/0000-0001-9901-8507}{\includegraphics[scale=0.06]{orcid.pdf}\hspace{1mm}Dries F.~Benoit} \\
	Ghent University\\
	CVAMO Core Lab, Flanders Make\\
	\texttt{dries.benoit@ugent.be} \\
}

\renewcommand{\headeright}{Published as a workshop paper at ECML-PKDD 2024}
\renewcommand{\undertitle}{Published as a workshop paper at ECML-PKDD 2024}
\renewcommand{\shorttitle}{}

\hypersetup{
	pdftitle={Active Learning to Guide Labeling Efforts for Question Difficulty Estimation},
	pdfsubject={cs.LG, cs.CL},
	pdfauthor={Arthur Thuy, Ekaterina Loginova, Dries F.~Benoit},
	pdfkeywords={Question Difficulty Estimation, Natural language processing, Active learning, Monte Carlo dropout, PowerVariance},
}

\begin{document}
\maketitle

\begin{abstract}
	In recent years, there has been a surge in research on Question Difficulty Estimation (QDE) using natural language processing techniques. 
	Transformer-based neural networks achieve state-of-the-art performance, primarily through supervised methods but with an isolated study in unsupervised learning. 
	While supervised methods focus on predictive performance, they require abundant labeled data. 
	On the other hand, unsupervised methods do not require labeled data but rely on a different evaluation metric that is also computationally expensive in practice.
	This work bridges the research gap by exploring active learning for QDE---a supervised human-in-the-loop approach striving to minimize the labeling efforts while matching the performance of state-of-the-art models.
	The active learning process iteratively trains on a labeled subset, acquiring labels from human experts only for the most informative unlabeled data points.
	Furthermore, we propose a novel acquisition function PowerVariance to add the most informative samples to the labeled set, a regression extension to the PowerBALD function popular in classification.
	We employ DistilBERT for QDE and identify informative samples by applying Monte Carlo dropout to capture epistemic uncertainty in unlabeled samples. 
	The experiments demonstrate that active learning with PowerVariance acquisition achieves a performance close to fully supervised models after labeling only 10\% of the training data.
	The proposed methodology promotes the responsible use of educational resources, makes QDE tools more accessible to course instructors, and is promising for other applications such as personalized support systems and question-answering tools.
\end{abstract}


% =====================================================================
% Main matter
% =====================================================================

% =====================================================================

% NOTE: use \citep and \citet

\section{Introduction}
\label{sec:intro}

Question Difficulty Estimation (QDE), also known as question calibration, is a regression task that estimates a question's difficulty directly from the question and answers' text.
It is a crucial task in personalized support tools like computerized adaptive testing \citep{van2000computerized}, which tailors questions to a student's skill level.
If the questions are too easy or too difficult, the student might lose motivation, negatively affecting their learning outcome \citep{wang2014regularized}.

Traditionally, QDE has been performed with manual calibration \citep{attali2014estimating} and pretesting \citep{lane2016handbook}, which are time-consuming and expensive.
Recent studies aim to address these limitations by leveraging natural language processing (NLP) techniques.
The NLP approaches train machine learning models to estimate question difficulty from its text.
Once trained, the models can quickly calibrate unseen questions, reducing the need for pretesting and manual calibration.

Supervised techniques dominate QDE with state-of-the-art results \citep{zhou2020multi,benedetto2021application} by fine-tuning the publicly available pre-trained models BERT \citep{devlin2018bert} and DistilBERT \citep{sanh2019distilbert}. 
However, fine-tuning often requires a large labeled dataset containing tens of thousands of calibrated questions, almost impossible to collect for individual course instructors developing QDE tools on their exam data. 
An isolated study \citep{loginova2021towards} has delved into an unsupervised approach, relying solely on additional pre-training and evaluating pairwise difficulty. 
Although this approach is helpful, its performance cannot be directly compared to supervised methods and is more computationally expensive in practical implementations.

In this work, we explore \textit{active learning (AL)} \citep{settles2009active} for QDE, a data-efficient supervised approach aiming to minimize the labeling work for human annotators while matching the performance of state-of-the-art models.
AL operates by iteratively training a model on an increasingly growing labeled subset by acquiring labels from an expert only for the most informative unlabeled data points.
This human-in-the-loop strategy allows us to preserve the well-established supervised evaluation methods, effectively bridging the gap between the performance-driven supervised domain and the data-centric unsupervised domain.
Moreover, we propose a novel acquisition function \textit{PowerVariance} to add the most informative samples to the labeled set while limiting redundant information, a regression extension to the PowerBALD function \citep{kirsch2021stochastic} popular in classification.
We use DistilBERT \citep{sanh2019distilbert} for QDE and find informative samples by applying Monte Carlo (MC) dropout \citep{gal2016dropout} to capture epistemic uncertainty over the unlabeled samples.

The proposed methodology contributes to the responsible use of educational resources by drastically reducing the labeling work, making the development of QDE tools more accessible to course instructors.
The findings have positive implications for a variety of applications like personalized support tools, essay correction tools, and question-answering systems.

The remainder of the paper is organized as follows. 
Section \ref{sec:related_work} provides an overview of related work, followed by the proposed AL methodology in Section \ref{sec:methodology}. 
Experimental details are discussed in Section \ref{sec:experiments}, with the results and discussion presented in Section \ref{sec:results_discussion}. 
Finally, Section \ref{sec:conclusion} concludes the paper. 
The code is available in a GitHub repository.\footnote{\url{https://github.com/arthur-thuy/qde-active-learning}}

% =====================================================================

\section{Related Work}
\label{sec:related_work}

Earliest NLP research on QDE from text primarily focused on multiple-choice questions (MCQs), employing bag-of-words techniques and assessing similarities among questions, correct choices, and incorrect choices \citep{alsubait2013similarity,yaneva2018automatic,kurdi2017experimental}.
However, these methods have been outperformed by more recent machine learning approaches.

Machine learning approaches to QDE fall into two main categories: (i) those involving distinct feature engineering and regression phases, and (ii) end-to-end neural networks (NNs).
The former encompasses a wide range of features, including linguistic features, text embeddings, frequency-based features, and readability indexes.
Several studies have also explored combinations of these feature techniques \citep{benedetto2023quantitative}. Common machine learning regression models in this group include random forests, support vector machines, and linear regression \citep{benedetto2023survey}.

End-to-end NN approaches in previous research primarily rely on Transformer models \citep{vaswani2017attention}, which can be either supervised or unsupervised.
Transformers are attention-based NNs pre-trained on a large corpus of text.
This generally yields superior performance with shorter training times compared to training NNs from scratch, leveraging the pre-existing knowledge of the pre-trained model.

Supervised estimation to QDE is most prevalent in the literature \citep{cheng2019dirt,qiu2019question,tong2020exercise}.
Fine-tuning the publicly available pre-trained models BERT \citep{devlin2018bert} and DistilBERT \citep{sanh2019distilbert} on the task of QDE gives state-of-the-art results \citep{zhou2020multi,benedetto2021application} and has been shown to outperform other approaches using traditional NLP-derived features \citep{benedetto2023quantitative}.

Unsupervised estimation, aiming to avoid relying on labeled data entirely, has received comparatively less attention. One study \citep{loginova2021towards} estimates question difficulty by leveraging the epistemic uncertainty in question answering models as an indicator of human-perceived difficulty.
This approach involves additional pre-training without fine-tuning, making it independent of labeled data.
While helpful in estimating difficulty, its performance cannot be directly compared to supervised estimation as it evaluates pairwise difficulty.
Moreover, it poses computational challenges in practice because numerous pairwise evaluations are required to determine an overall difficulty ranking of unseen questions.

% =====================================================================

\section{Methodology}
\label{sec:methodology}

\subsection{Active Learning}

\begin{figure}[t!]
	\centering
	\includegraphics[width=0.7\textwidth]{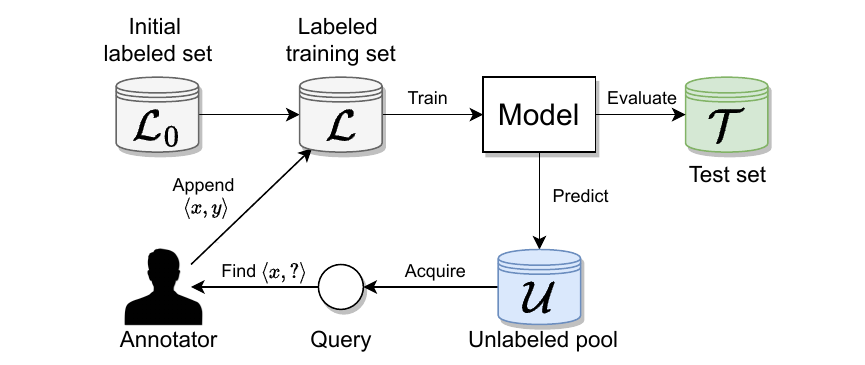} 
	\caption{Active learning workflow with pool-based sampling. Active learning iteratively trains on a subset of labeled data and acquires labels from an expert annotator for samples in the unlabeled pool. Adapted from~\citep{settles2009active}.}
	\label{fig:al_workflow}
\end{figure}

AL \citep{settles2009active} is a human-in-the-loop technique for achieving data efficiency.
Instead of collecting and labeling a large dataset before training, which is time-consuming and expensive, AL iteratively acquires labels from an expert annotator only for the most informative data points from a pool of unlabeled data.
After each acquisition step, the newly labeled points are added to the training set, and the model is retrained.
This process is repeated until reaching a desired level of accuracy or until the labeling budget is exhausted, aiming to minimize the labeling work of human annotators.
Figure \ref{fig:al_workflow} provides a visual overview of the AL workflow, employing pool-based sampling as described.

In AL, the informativeness of new points is assessed by an acquisition function.
The acquisition function typically relies on epistemic uncertainty over unlabeled data, which can be obtained with approximate Bayesian inference techniques like MC dropout \citep{gal2016dropout} or with ensembling techniques \citep{lakshminarayanan2017simple,thuy2023explainability,thuy2024reliable}.
Epistemic uncertainty represents uncertainty in the model parameters and is naturally high in regions of the input space with few training observations \citep{der2009aleatory}, precisely the observations we want to add to the labeled set.
For classification tasks, a commonly used acquisition scoring function is Bayesian Active Learning by Disagreement (BALD) \citep{houlsby2011bayesian}, estimates the epistemic uncertainty by measuring the variability among samples of the predictive distribution. 
Data points maximize this acquisition function when the model assigns the highest predicted probability to a different class in each sample.
For regression tasks, the epistemic uncertainty is estimated by the Variance among predictive samples \citep{settles2009active}.
Similarly, data points score high on this acquisition function when the model’s output varies strongly across the samples.

\subsection{Monte Carlo Dropout Uncertainty}

Uncertainty in predictions can arise from two different sources: aleatoric and epistemic uncertainty \citep{der2009aleatory}.
Aleatoric uncertainty refers to the notion of randomness and is related to the data-measurement process. 
This uncertainty is irreducible even if more data is collected.
Epistemic uncertainty accounts for uncertainty in the model parameters. 
In contrast to data uncertainty, collecting more data can reduce model uncertainty.
As such, it is interesting for acquisitions functions to select the unlabeled samples with the largest epistemic uncertainty.

We assume a regression task with inputs $\mathbf{X}$, labels $\mathbf{Y}$, and a discriminative regressor $p(\mathbf{y} \mid \mathbf{x})$.
For the Bayesian MC dropout models, we further assume a probability distribution over the parameters, $p(\bm{\theta})$, and we have $p(\mathbf{y} \mid \mathbf{x}) = \EX_{p(\bm{\theta})} [p(\mathbf{y} \mid \mathbf{x},\ \bm{\theta})]$.
In a NN regressor, the output $\mathbf{y}$ represents the mean $\mu_{\mathbf{x}}$ of the conditional probability distribution $\mathcal{N}(\mu_{\mathbf{x}}, \sigma=1)$, for some input point $\mathbf{x}$.
The standard NN regressor only outputs a single $\mu_{\mathbf{x}}$, hence does not capture any uncertainty.
With MC dropout, multiple estimates for $\mu_{\mathbf{x}}$ are obtained and the variance over these estimates is an approximation for the epistemic uncertainty in data point $\mathbf{x}$. 

In MC dropout \citep{gal2016dropout}, dropout is not only applied at training time but also at test time. 
Multiple forward passes are performed, each time randomly dropping units and getting another thinned dropout variant of the NN.
As such, it can be seen as an implicit ensemble method where each sample corresponds to an ensemble member.
The various samples approximate the true posterior predictive distribution, enabling it to estimate the epistemic uncertainty in a data point.

\subsection{PowerVariance Acquisition}
\label{subsec:powervariance}

\begin{figure}[t!]
	\centering
	\includegraphics[width=0.7\textwidth]{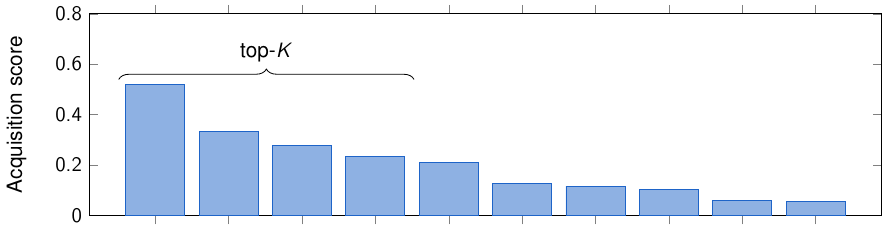}
	\caption{Top-$K$ acquisition toy example. Acquisition scores for each unlabeled pool point are ordered and the top-$K$ points are selected.}
	\label{fig:topk_acquisition}
\end{figure}

In practical AL applications, instead of single data points, batches of data points are selected in each acquisition step to minimize the frequency of model retraining and expert involvement.
A common heuristic involves selecting the top-$K$ highest-scoring points from an acquisition scheme designed for single-point selection, i.e., top-$K$ acquisition \citep{kirsch2019batchbald} (Figure \ref{fig:topk_acquisition}).
However, this method overlooks interactions between points within an acquisition batch because individual points are scored independently.
For example, if the most informative point is duplicated in the pool set, all instances will be acquired, which is wasteful.

To address this issue, acquisition functions designed explicitly for batch acquisition with NN classifiers have been developed, such as BatchBALD \citep{kirsch2019batchbald}. 
These methods improve over top-$K$ acquisition by accounting for the interaction between points but are computationally expensive. 
To limit the computational burden, the authors of \citet{kirsch2021stochastic} propose to stochastically acquire points following a power distribution determined by the single-acquisition scores.
Intuitively, points with high acquisition scores are more likely to be sampled.
For example, for BALD, the method is referred to as PowerBALD, demonstrating equal performance to state-of-the-art batch acquisition functions like BatchBALD while requiring significantly less computational resources.

The stochastic acquisition strategy \citep{kirsch2021stochastic} assumes that as new samples are selected in a batch, future acquisition scores differ from the current scores by a perturbation. This perturbation is modeled as Gumbel-distributed noise for two reasons.

First, to select the $k$-th point in the acquisition batch of size $K$, it is important to consider how much additional information (i.e., increase in acquisition scores) the still-to-be-selected $K - k$ points will provide.
As such, the stochastic strategy models the maximum future increase in acquisition scores over all possible candidate points to complete the batch.
Empirically, acquisition scores are similar to a truncated exponential distribution, with different rate parameters at each AL step.
The maximum over sums of such random variables is empirically shown to follow a Gumbel distribution \citep{kirsch2021stochastic}.

Second, the Gumbel distribution is also mathematically convenient.
The Gumbel-Top-$K$ trick \citep{kool2019stochastic} shows that taking the highest-scoring points from a distribution perturbed with Gumbel noise is equivalent to sampling from a softmax distribution without replacement. 
Building on this, perturbing the log-scores with Gumbel noise results in sampling from a power distribution. 
Power acquisition assumes that scores are non-negative and uninformative points should be avoided, a sensible approach for AL.

We propose to extend this approach to regression settings, which is currently underinvestigated, resulting in a PowerVariance acquisition function.
Similar to BALD, the Variance scoring function is non-negative, with zero variance indicating an uninformative sample due to no expected information gain. 
Consequently, the Variance function should also couple well with power acquisition, mirroring the success seen with BALD and PowerBALD.

More formally, for each candidate pool index $i$, the Variance score is
\begin{equation}
	s_{Var}(i) = \Var[p(\mathbf{y} \mid \mathbf{x}_i,\ \bm{\theta})].
\end{equation}
The PowerVariance score is the perturbation of the log Variance score with Gumbel-distributed noise $\epsilon_i \sim \textrm{Gumbel}(0;\beta^{-1})$
\begin{equation}
	s_{PowerVar}(i) = \log s_{Var}(i) + \epsilon_i.
\end{equation}
Following the Gumbel-Top-$K$ trick \citep{kool2019stochastic}, taking the top-$K$ points from $s_{PowerVar}$ is equivalent to sampling without replacement from the distribution $p_{PowerVar}$ 
\begin{equation}
	p_{PowerVar}(i) \propto s_{Var}(i)^{\beta}
\end{equation}
where $\beta \geq 0$ is a \textit{coldness} parameter.
Note that the coldness parameter $\beta$ is different but similar to the \textit{temperature} parameter $T = 1/\beta$ often used in text-generation with language models.
For $\beta \rightarrow \infty$, this strategy converges towards top-$K$ acquisition as it is more likely to only sample points with a high score. 
For $\beta \rightarrow 0$, it converges towards uniform acquisition because it is more likely to also sample points with a low score.

% =====================================================================

\section{Experiments}
\label{sec:experiments}

\subsection{Data}

RACE\texttt{++} \citep{liang2019new} is a dataset of reading comprehension MCQs, built by merging the original RACE dataset \citep{lai2017race} with the RACE-C dataset \citep{liang2019new}.
Each question comprises a reading passage, a stem, and four possible answer options, one of them being correct.
Each question has one out of three difficulty levels (0, 1, 2), which we consider as the gold standard for QDE.
The difficulty levels correspond to middle school, high school, and university-level questions; the dataset is imbalanced, with a distribution of 25\%, 62\%, and 13\% respectively.
Note that the dataset labels are all available; the labels are hidden and revealed once requested by the acquisition function.
The training split contains \num{100568} questions, while the validation and test splits contain \num{1000} and \num{5642} questions, respectively.
There are no reading passages shared across the splits.

\subsection{Model Architecture}

We fine-tune the publicly available pre-trained model DistilBERT on the task of QDE. 
DistilBERT is a language model obtained by distilling BERT, i.e., compressing BERT by training a small model to reproduce its full output distribution \citep{hinton2015distilling}.
The authors of \citet{benedetto2023survey} find that DistilBERT achieves comparable performance to BERT on QDE while using approximately half the parameter count.
Limiting the computational expense is important in AL as the model needs to be fine-tuned over multiple iterations.

To adapt DistilBERT for QDE, we stack a fully connected hidden layer on top of the pre-trained language model, followed by the output layer.
The regression output layer has one unit, with its weights initialized randomly.
During fine-tuning, both the weights of the output head and the pre-trained model are updated.
We follow the input encoding of \citet{benedetto2023quantitative} and concatenate the question and the text of all the possible answer choices in a single sentence, divided by separator tokens.
This configuration has demonstrated slight improvements over using no answer choice at all or only the correct answer.

Following previous research \citep{benedetto2023quantitative}, we handle QDE on the RACE\texttt{++} dataset as a discrete regression problem.
The QDE model is trained as a regression model and outputs a continuous difficulty, which is then converted to the closest discrete level with simple thresholds.
As evaluation metric, we compute the root mean squared error (RMSE) between the discrete predictions and discrete difficulty levels because of its consideration for the order of difficulty levels.
We refer to this metric as ``discrete RMSE''.

\subsection{Active Learning Setup}

The AL process starts with an initial labeled dataset of 500 observations, randomly selected from the training set and following the training set distribution (i.e., 25\%/62\%/13\%).
In each iteration, the model is fine-tuned for 10 training epochs and the parameters giving the best validation performance are saved.
We use the AdamW optimizer \citep{loshchilov2017decoupled} with learning rate 2e-5 and batch size 64.

Subsequently, the model is evaluated on a random subset of the unlabeled pool containing \num{5000} observations, from which \num{100} observations are selected for labeling and added to the training set.
As demonstrated by \citet{atighehchian2019baal}, using a random subset instead of the entire pool minimally impacts predictive performance while being more computationally efficient.
The configurations with MC dropout use 10 MC samples to calculate epistemic uncertainty.
Before training on the new training set, the model weights are re-initialized to their original state.
This involves using pre-trained weights for the base model and randomly initialized weights for the output layer with He initialization \citep{he2015delving}, effectively mitigating the risk of getting stuck in a poor local minimum.
Table \ref{tab:hyperparams} displays all hyperparameter settings for the model and AL setup.

\begin{table}[t!]
	\caption{Hyperparameter settings}
	\label{tab:hyperparams}
	\begin{subtable}[t]{.5\linewidth}
		\centering
		\caption{Model}
		\begin{tabular}{lc}  % NOTE: if less space, do "{@{}lccc@{}}"
			\toprule%
			\textbf{Hyperparameter} & \textbf{Value} \\
			\midrule
			Model name & DistilBERT \\
			Learning rate & 2e-5 \\
			Optimizer & AdamW \\
			Weight decay & 0.05 \\
			Loss function & MSE \\
			Training epochs & 10 \\
			Train batch size & 64 \\
			Eval batch size & 256 \\
			Dropout rate & 0.1 \\
			Warmup ratio & 0.1 \\
			Sequence length & 256 \\
			\bottomrule
		\end{tabular}
	\end{subtable}%
	\begin{subtable}[t]{.5\linewidth}
		\centering
		\caption{Active learning}
		\begin{tabular}{lc}  % NOTE: if less space, do "{@{}lccc@{}}"
			\toprule%
			\textbf{Hyperparameter} & \textbf{Value} \\
			\midrule
			Dataset name & RACE\texttt{++} \\
			\makecell[bl]{Data size \\ train/val/test} & \num{100568}/\num{1000}/\num{5642} \\
			Initial labeled set size & \num{500} \\
			Acquisition size & \num{100} \\
			Pool subset size & \num{5000} \\
			Final labeled set size & \num{10000} \\
			MC samples & 10 \\
			\bottomrule
		\end{tabular}
	\end{subtable} 
\end{table}

In the experiments, we compare three AL configurations: (i) Uniform acquisition with a standard NN, (ii) top-$K$ Variance acquisition with an MC dropout NN, and (iii) PowerVariance acquisition with an MC dropout NN.
For PowerVariance acquisition, we follow \citet{kirsch2021stochastic} and set $\beta=1$ to limit the number of hyperparameters.
Note that Uniform acquisition is computationally cheaper because it does not predict on the unlabeled pool, instead it randomly selects observations for labeling.

Additionally, we investigate the performance of three baselines: (i) Random, (ii) Majority, and (iii) Supervised. 
The Random baseline randomly predicts a difficulty level, the Majority baseline consistently predicts level 1 (the most prevalent level in the training set), and the Supervised baseline fine-tunes a model on the fully labeled training set. 
Consequently, the Random and Majority baselines serve as performance lower bounds, while the Supervised baseline sets an upper bound.

The experiments are implemented in PyTorch \citep{Ansel_PyTorch_2_Faster_2024} using the BaaL~\citep{atighehchian2019baal} and HuggingFace~\citep{wolf2020transformers} packages, executed on an NVIDIA RTX A5000 GPU. The results are averaged over five independent runs with random seeds, with a total runtime of 90 hours.

% =====================================================================

\section{Results and Discussion}
\label{sec:results_discussion}

Section \ref{subsec:pred_performance} explores the AL results, while Section \ref{subsec:acquisitions} provides a more detailed analysis of the behavior of the acquisition functions.

\subsection{Predictive Performance}
\label{subsec:pred_performance}

\begin{figure}[t!]
	\centering
	\includegraphics[width=0.6\textwidth]{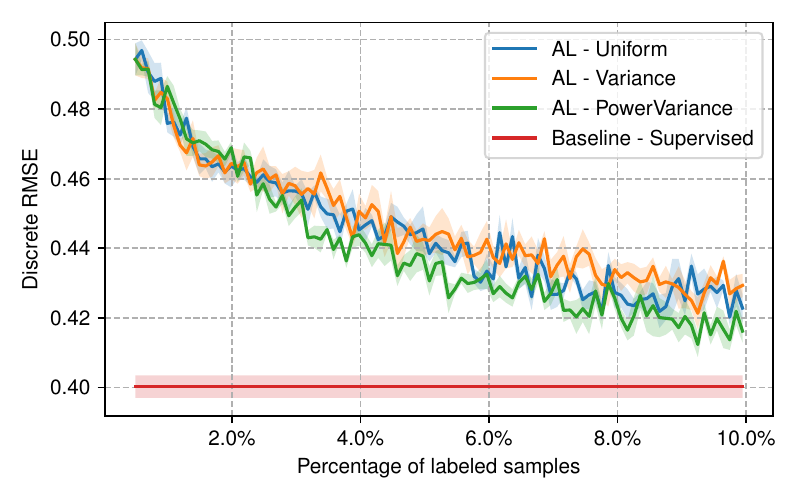} 
	\caption{Discrete RMSE as a function of the labeled dataset size. PowerVariance acquisition outperforms Uniform and Variance acquisition by achieving the lowest discrete RMSE scores as AL progresses. After labeling 10\% of the data, its performance is close to the fully supervised model.}
	\label{fig:metric_vs_dataset}
\end{figure}

Figure \ref{fig:metric_vs_dataset} shows the discrete RMSE in relation to the training dataset size.
For AL configurations, values to the lower left indicate better performance.
The Supervised baseline's performance is represented by a horizontal line, where lower is better.
The Random baseline achieves a discrete RMSE of \num{1.026} and the Majority baseline achieves \num{0.616}.
Note that these baselines are not shown in Figure \ref{fig:metric_vs_dataset} to avoid an excessively large vertical axis, which complicates interpretation.

The findings reveal that fine-tuning on the initial labeled set (500 observations; 0.5\%) performs exactly in between the Majority and Supervised baselines. 
As the labeled training set expands, we observe a decrease in discrete RMSE across all AL configurations.
This finding demonstrates that DistilBERT performs well with limited labeled data, consistent with previous research \citep{sun2019fine} using the non-distilled BERT model.

Variance acquisition disappoints and performs on par with Uniform acquisition.
This result is surprising given that Variance acquisition uses MC dropout to quantify epistemic uncertainty over the unlabeled pool points.
As such, naively selecting the top-$K$ highest scoring-points does not yield improved results.

In contrast, PowerVariance acquisition outperforms both Uniform and Variance acquisition, achieving the lowest discrete RMSE score from 2\% of labeled samples onwards.
Although PowerVariance's RMSE advantage over Random acquisition appears minimal due to the line curves being close, a substantial number of labeled questions is needed to overcome this advantage.
After collecting a labeled set containing 10\% of the available samples, PowerVariance reaches a discrete RMSE score only 5\% higher than training on 100\% of the training data.

\begin{figure}[t!]
	\centering
	\includegraphics[width=0.6\textwidth]{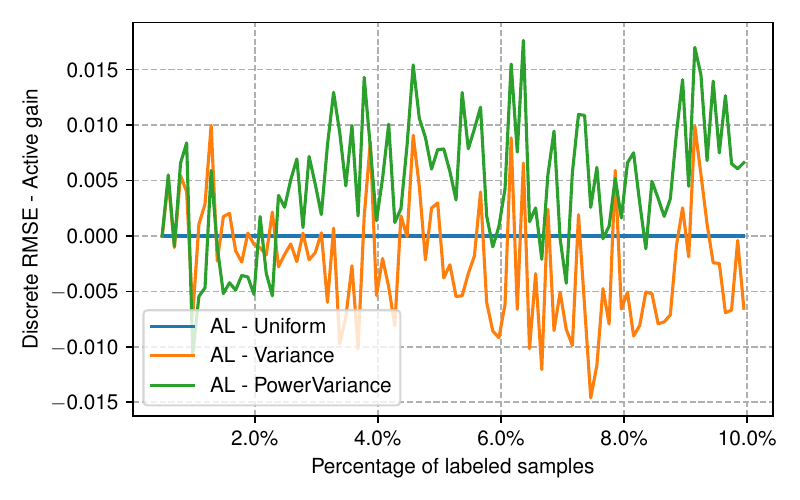} 
	\caption{Active gain over Uniform acquisition as a function of the labeled dataset size. Variance acquisition performs on par with passive learning, while PowerVariance offers an active gain of 0.01 discrete RMSE.}
	\label{fig:active_gain}
\end{figure}

AL with Uniform acquisition is often referred to as \textit{passive learning}, as samples are randomly selected from the unlabeled pool.
Figure \ref{fig:active_gain} illustrates the active gain in discrete RMSE, highlighting the performance differences of Variance and PowerVariance over Uniform acquisition.
Positive values denote an advantage, while negative values indicate a disadvantage, enabling relative comparisons among acquisition functions.

From 2\% of labeled data onwards, PowerVariance exhibits a positive active gain, averaging around 0.01 discrete RMSE. 
In contrast, Variance acquisition does not offer advantages over passive learning. The next subsection delves deeper into the acquisition functions, examining the reasons behind the performance differences.

\subsection{Acquisition Behavior}
\label{subsec:acquisitions}

\begin{figure}[t!]
	\centering
	\begin{subfigure}{.45\linewidth}
		\centering
		\includegraphics[width=1.0\linewidth]{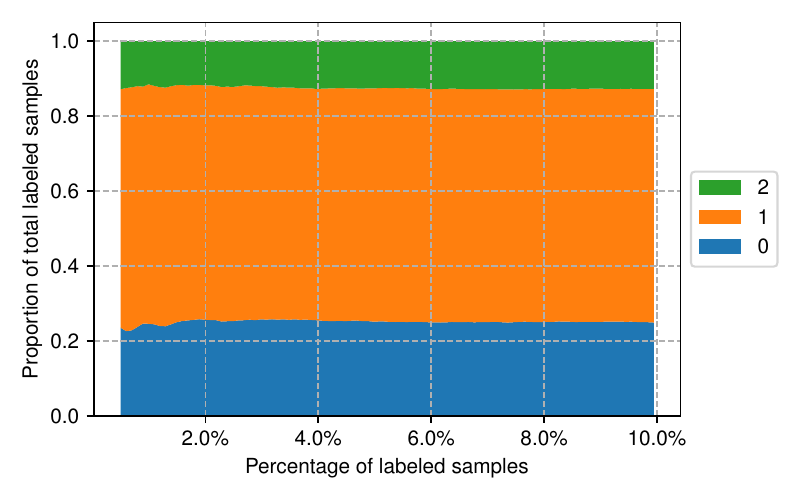}
		\caption{Uniform}
		\label{fig:uniform_acquisition}
	\end{subfigure}%
	\begin{subfigure}{.45\linewidth}
		\centering
		\includegraphics[width=1.0\linewidth]{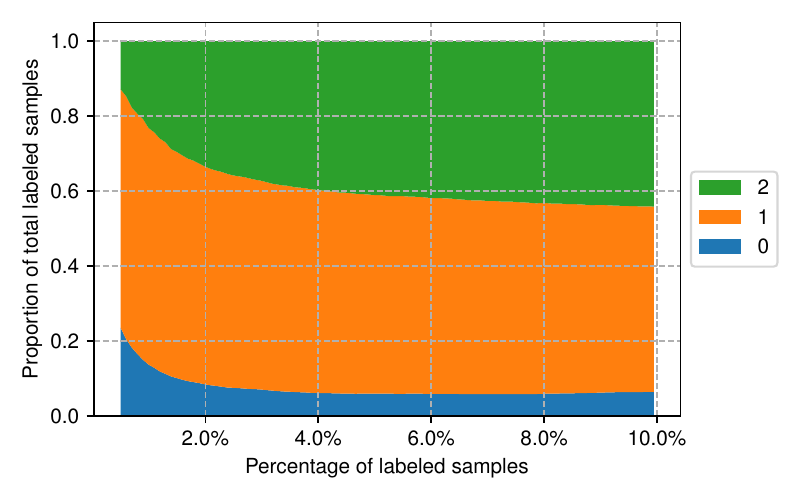}
		\caption{Variance}
		\label{fig:variance_acquisition}
	\end{subfigure}\\[1ex]
	\begin{subfigure}{.45\linewidth}
		\centering
		\includegraphics[width=1.0\linewidth]{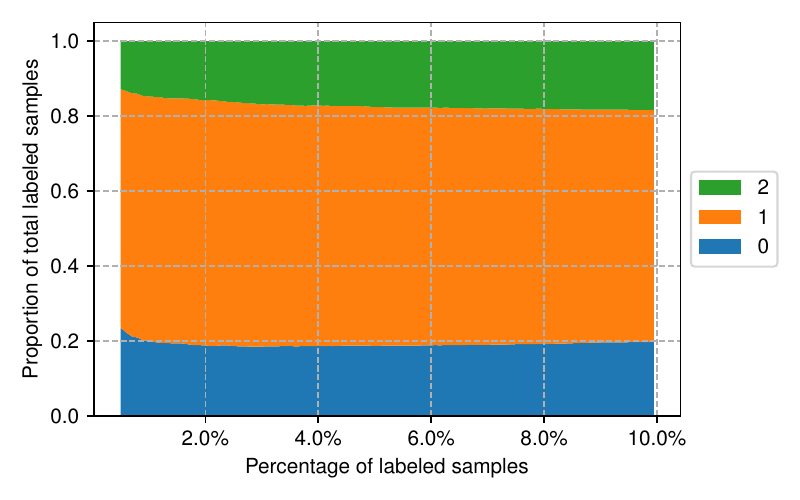}
		\caption{PowerVariance}
		\label{fig:powervariance_acquisition}
	\end{subfigure}%
	\caption{Distribution of difficulty levels in the labeled set as a function of the labeled dataset size, per acquisition function. Similar to Variance, PowerVariance selects more level 2 observations but does not neglect level 0 samples.}
	\label{fig:acquisitions}
\end{figure}

To better understand how the acquisition functions behave, we visualize the distribution of difficulty levels in the labeled set as training progresses (see Figure \ref{fig:acquisitions}). 
Initially, the labeled set is randomly sampled from the pool following a 25\%/62\%/13\% distribution for levels 0, 1, and 2, respectively.
Due to the small sample size (500 samples), slight deviations from this distribution are possible.

As expected, Uniform acquisition causes minimal changes in the level distribution because samples are randomly selected from the unlabeled pool.
In contrast, Variance acquisition exhibits a distinctive pattern, selecting many level 2 observations and few level 0 instances.
The proportion of level 2 samples increases from 13\% to 45\%, while level 0 samples decrease from 25\% to merely 6\%.
These findings partially align with previous studies \citep{atighehchian2020bayesian} suggesting that top-$K$ strategies using epistemic uncertainty target underrepresented classes.
Variance acquisition indeed prioritizes sampling from the most underrepresented class (level 2) but does this primarily at the expense of level 0 observations, rather than the majority class (level 1).

PowerVariance exhibits behavior that falls between Uniform and Variance acquisition. 
Like Variance acquisition, it selects more level 2 observations, increasing their proportion from 13\% to 19\%, while only slightly reducing the level 0 proportion from 25\% to 20\%. 
As such, it is a less aggressive approach compared to Variance acquisition.

\begin{figure}[t!]
	\centering
	\begin{subfigure}{.45\linewidth}
		\centering
		\includegraphics[width=1.0\linewidth]{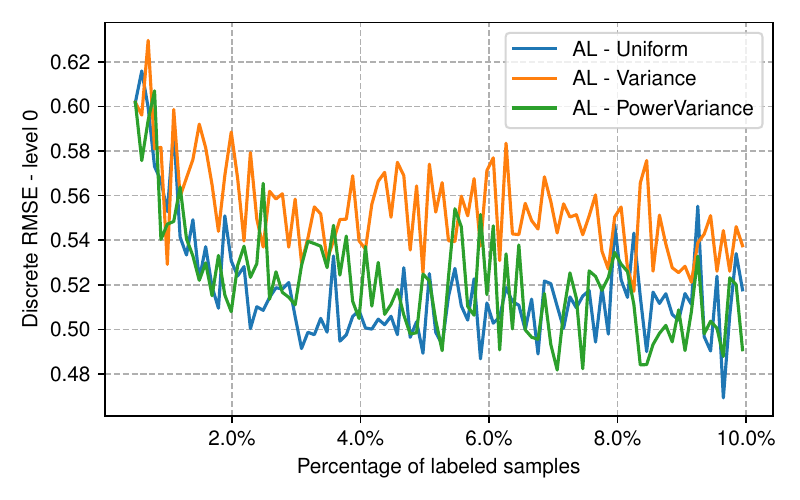}
		\caption{Level 0 (middle school)}
		\label{fig:level_0}
	\end{subfigure}%
	\begin{subfigure}{.45\linewidth}
		\centering
		\includegraphics[width=1.0\linewidth]{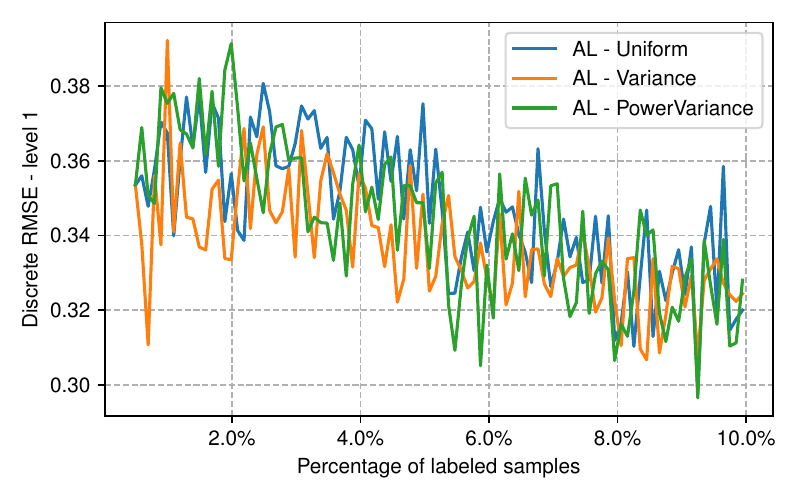}
		\caption{Level 1 (high school)}
		\label{fig:level_1}
	\end{subfigure}\\[1ex]
	\begin{subfigure}{.45\linewidth}
		\centering
		\includegraphics[width=1.0\linewidth]{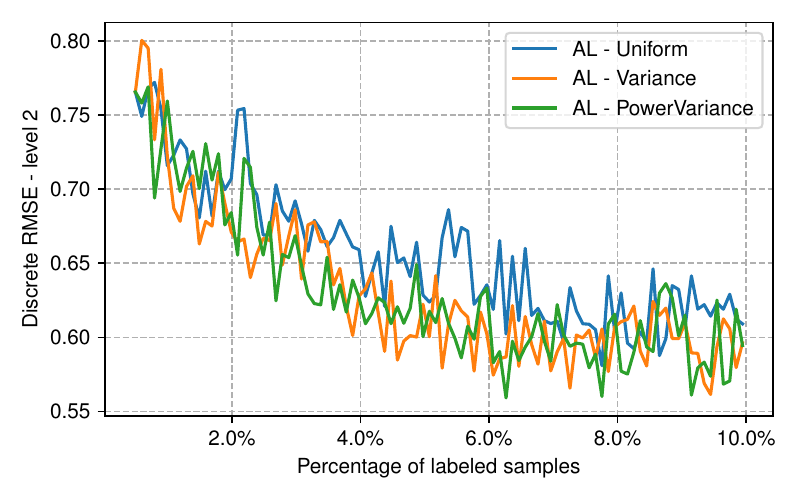}
		\caption{Level 2 (university)}
		\label{fig:level_2}
	\end{subfigure}%
	\caption{Discrete RMSE as a function of the dataset size, per individual difficulty level. Variance and PowerVariance surpass Uniform acquisition on level 2, but Variance underperforms on level 0.}
	\label{fig:level_performance}
\end{figure}

Moreover, we analyze the impact of the acquisition strategies on the predictive performance for each difficulty level individually.
Figure \ref{fig:level_performance} displays the discrete RMSE performance per difficulty level.

For level 1 questions (Figure \ref{fig:level_1}), all acquisition functions have comparable performance as the line graphs overlap.
However, we observe performance differences on level 0 and level 2 questions.

For level 0 questions (Figure \ref{fig:level_0}), Variance acquisition performs poorly as the orange curve is notably higher than the other curves.
This poor performance is a direct consequence of neglecting level 0 observations during acquisition.

For level 2 questions (Figure \ref{fig:level_2}), Uniform acquisition performs worst.
Level 2 observations are the most difficult to estimate and the most underrepresented in the initial labeled set.
Variance and PowerVariance sample a large number of these questions and therefore achieve good RMSE scores, lower than Uniform acquisition.
It is also worth noting that the performance of Variance and PowerVariance is very similar, although Variance samples a much higher proportion of level 2 questions (45\%) than PowerVariance (19\%).

For each difficulty level, PowerVariance acquisition performs on par or better than Uniform and Variance acquisition.
It leverages epistemic uncertainty to sample more from underrepresented level 2 questions which are most challenging to estimate, whereas Uniform acquisition naively samples at random.
Furthermore, PowerVariance recognizes redundant uncertainty information in level 2 questions and instead samples from level 0 questions, whereas Variance neglects level 0 questions, significantly hampering its performance.

% =====================================================================

\section{Conclusion}
\label{sec:conclusion}

This work explores AL for QDE, a supervised approach aiming to minimize the labeling effort for human annotators while matching the performance of state-of-the-art models.
By using a human-in-the-loop method, it bridges the gap between the performance-driven supervised domain and the data-centric unsupervised domain.
Additionally, we introduce a novel acquisition function PowerVariance, which leverages epistemic uncertainty from unlabeled samples obtained through MC dropout to identify the most informative data points.
Unlike conventional Variance acquisition, PowerVariance is designed to limit redundant information in a batch of samples.

Experimental results indicate that the proposed PowerVariance acquisition outperforms both Uniform and Variance acquisition. 
It effectively selects observations from the minority difficulty level 2 for labeling and does not neglect level 0 questions, an issue observed with Variance acquisition.
We see no reason for practitioners to consider the flawed top-$K$ Variance acquisition.
Even with only 10\% of the training data labeled, AL with PowerVariance achieves good performance, only 5\% higher discrete RMSE than the model trained on 100\% of the training data.

This methodology promotes the responsible use of educational resources by significantly reducing the labeling work for educational professionals while maintaining predictive performance.
Consequently, it makes QDE tools more accessible to course instructors who might otherwise be demotivated by the large number of calibrated questions required.

The study is potentially limited by the small number of coarse difficulty levels. Course instructors are often reluctant to share exam questions, making it challenging to find datasets with more realistic difficulty levels. Future research can explore more fine-grained settings with more closely spaced difficulty levels. The inability to use public datasets highlights the relevance of active learning strategies for course instructors when labeling exam questions. Furthermore, adding difficulty levels may introduce class imbalance, a scenario where PowerVariance performs strongly.

The proposed AL approach holds promise for diverse applications such as personalized support tools, essay correction tools, and question-answering systems.
It can easily be adapted to alternative pre-trained language models and datasets, as MC dropout works on any architecture that uses dropout.
For models not employing dropout, ensembles of NNs can provide epistemic uncertainty, enabling similar AL strategies.

\section*{Acknowledgments}

This study was supported by the Research Foundation Flanders (FWO) (grant number 1S97022N).

% =====================================================================
% End matter
% =====================================================================

%\bibliographystyle{unsrtnat}
%\bibliography{references.bib}  %%% Uncomment this line and comment out the ``thebibliography'' section below to use the external .bib file (using bibtex) .

% Uncomment this section and comment out the \bibliography{references} line above to use inline references.

\end{document}